\documentclass[10pt,twocolumn,letterpaper]{article}

\usepackage{iccv}
\usepackage{times}
\usepackage{epsfig}
\usepackage{graphicx}
\usepackage{amsmath}
\usepackage{amssymb}
\usepackage{bm}
\usepackage{multirow}
\usepackage{algorithm}
\usepackage{algorithmicx}
\usepackage{algpseudocode}
\usepackage{makecell}
\usepackage{textcomp}

\usepackage[breaklinks=true,bookmarks=false]{hyperref}

\iccvfinalcopy 

\def\iccvPaperID{6019} 

\ificcvfinal\fi

\begin{document}

\title{Learn to Cluster Faces via Pairwise Classification}

\author{
    Junfu Liu\thanks{Co-first authors.~  liujeff97@gmail.com,~  qiudi@meituan.com},~~~
    Di Qiu\footnotemark[1],~~~
    Pengfei Yan\thanks{Corresponding author.~ yanpengfei03@meituan.com},~~~
    Xiaolin Wei\\
    Meituan\\
}
\maketitle
\ificcvfinal\thispagestyle{empty}\fi

\begin{abstract}
Face clustering plays an essential role in exploiting massive unlabeled face data. Recently, graph-based face clustering methods are getting popular for their satisfying performances. However, they usually suffer from excessive memory consumption especially on large-scale graphs, and rely on empirical thresholds to determine the connectivities between samples in inference, which restricts their applications in various real-world scenes. To address such problems, in this paper, we explore face clustering from the pairwise angle. Specifically, we formulate the face clustering task as a pairwise relationship classification task, avoiding the memory-consuming learning on large-scale graphs. The classifier can directly determine the relationship between samples and is enhanced by taking advantage of the contextual information. Moreover, to further facilitate the efficiency of our method, we propose a rank-weighted density to guide the selection of pairs sent to the classifier. Experimental results demonstrate that our method achieves state-of-the-art performances on several public clustering benchmarks at the fastest speed and shows a great advantage in comparison with graph-based clustering methods on memory consumption.
\end{abstract}

\section{Introduction}

Massive face data are accessible on the Internet nowadays, thus boosting the development of face analysis technology \cite{{guo2016ms}, {nech2017level}}. However, the large-scale unlabeled face data has resulted in quite high annotation prices, and human annotations are not always reliable. Therefore, exploiting unlabeled face data has attracted great interest. Aiming at assigning pseudo labels to unlabeled face images, face clustering is a fundamental face analysis task and has wide applications in real-world scenarios like the dataset preparation or cleaning for face recognition \cite{{schroff2015facenet}, {zhang2018accelerated}} and photo albums management \cite{zhu2011rank}.

\begin{figure}[t]
\begin{center}
   \includegraphics[width=0.9\linewidth]{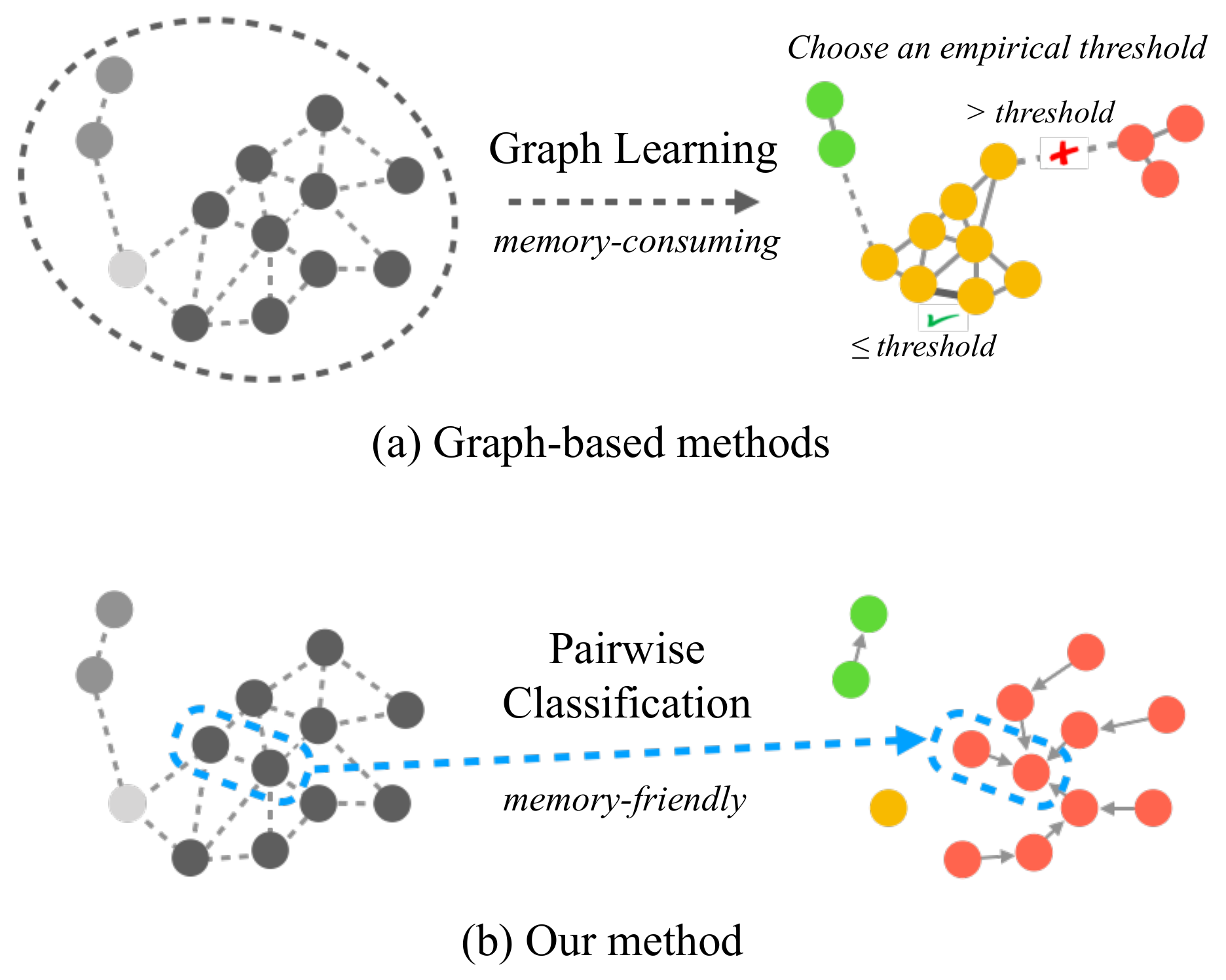}
\end{center}
   \caption{An overview of the main differences between popular graph-based face clustering methods and ours. The nodes with different colors represent different classes. (a) Graph-based methods focus on the learning of discriminative embeddings but need to set an empirical threshold to connect or disconnect samples. (b) Our method can directly determine the pairwise relationship through a classifier and generate the final clusters. Besides, by dealing with the face clustering task from graph-level to pair-level, our method largely reduces the memory consumption.}
\label{fig0}
\end{figure}

Traditional clustering methods suffer from impractical assumptions and their performances could be sensitive to hyper-parameters. For instance, DBSCAN \cite{ester1996density} assumes clusters to be in the same density, and the performance of K-Means \cite{lloyd1982least} highly depends on the original {\em k}. Recently, graph-based face clustering methods have drawn much attention for their satisfying performances by learning representative embeddings without making oversimplified assumptions \cite{{wang2019linkage}, {yang2019learning}, {zhan2018consensus}}. But the computational costs and demands for memory usage stand out especially on large-scale face clustering scenes, due to the surprisingly high price of learning on large graphs \cite{{kipf2017semi}, {chen2018fastgcn}}. And some works have been proposed to alleviate such problems. In \cite{yang2020learning}, Yang {\em et al.} designed a graph convolutional network-based confidence estimator to select nodes with large confidence, which reduces the number of highly overlapped subgraphs and leads to promotions on efficiency. Although the confidence estimator contains only one single layer, an out-of-memory issue still occurs when dealing with large-scale graphs, as will be revealed in Section \ref{sec4}. Guo {\em et al.} \cite{guo2020density} exploited cluster-level feature distribution by defining a density-aware graph, which reduces memory consumption by avoiding learning on the whole graph, but this method can still fall on large-scale face clustering scenes when limited memory capacity is available. More importantly, like previous graph-based clustering methods, to generate the final clusters in the inference stage, it still requires a manual-setting threshold to determine the connectivities between samples. The final performance of these clustering methods is so sensitive to the empirical threshold that can drop heavily when the threshold is set improperly, which restricts the generalization of various real-world scenes.

To address the problems above, in this work, we focus on the fundamental concerns of face clustering tasks, {\em i.e.}, the homogeneity of samples. Specifically, we explore a memory-friendly and threshold-free face clustering method from the pairwise angle, leading to promotions in both efficiency and accuracy. Instead of learning on the graph-level or cluster-level, we handle face clustering on the minimum level, {\em i.e.}, the pair-level, which largely reduces memory consumption. We adopt a simple yet effective classifier network and the classifier directly gives the connection relationship of a pair of samples, which also frees us from the manual setting of thresholds. Besides, inspired by the graph learning-based methods where contextual information is emphasized, the classifier is trained on our designed weighted-neighbor features, which brings considerable performance gains to both the classifier and the overall clustering task. As Figure~\ref{fig0} illustrates, popular graph-based methods rely on an empirical threshold to determine the connectivities between samples after learning distinguishable feature embeddings with high consumption on memory. Contrastively, our method can directly give the pairwise relationships and generate the final clusters efficiently with very limited memory usage.

Moreover, to enhance the efficiency of the whole pipeline, we propose a rank-weighted density. The rank-weighted density gives more attention to nearby neighbors, which eases the influences of outliers in the procedure of finding neighbors with higher density. The introduction of this more precise rank-weighted density not only reduces the time complexity via selecting fewer pairs sent to the classifier by an order of magnitude, but also makes a further contribution to the performance of our method.

Our main contributions are summarized as follows: (1) Different from popular graph-based methods, we propose a face clustering method based on pairwise classification, which is much more memory-friendly and also threshold-free. (2) A rank-weighted density is proposed to instruct the pair selection sent to the classifier, resulting in further promotions in both efficiency and accuracy. (3) Our method improves the performance on public face clustering benchmarks, achieving Pairwise F-score at 90.67 and BCubed F-score at 89.54 on the MS-Celeb-1M dataset, surpassing previous public state-of-the-art methods at 90.60 \cite{guo2020density} and 86.09 \cite{yang2020learning} respectively.

The following Section~\ref{sec2} will give a brief review of the related work. Section~\ref{sec3} details our proposed method. Experimental results and analysis are presented in Section~\ref{sec4} and we finally conclude the paper in Section~\ref{sec5}.

\section{Related Work}
\label{sec2}
\begin{figure*}
\begin{center}
\includegraphics[width=0.85\linewidth]{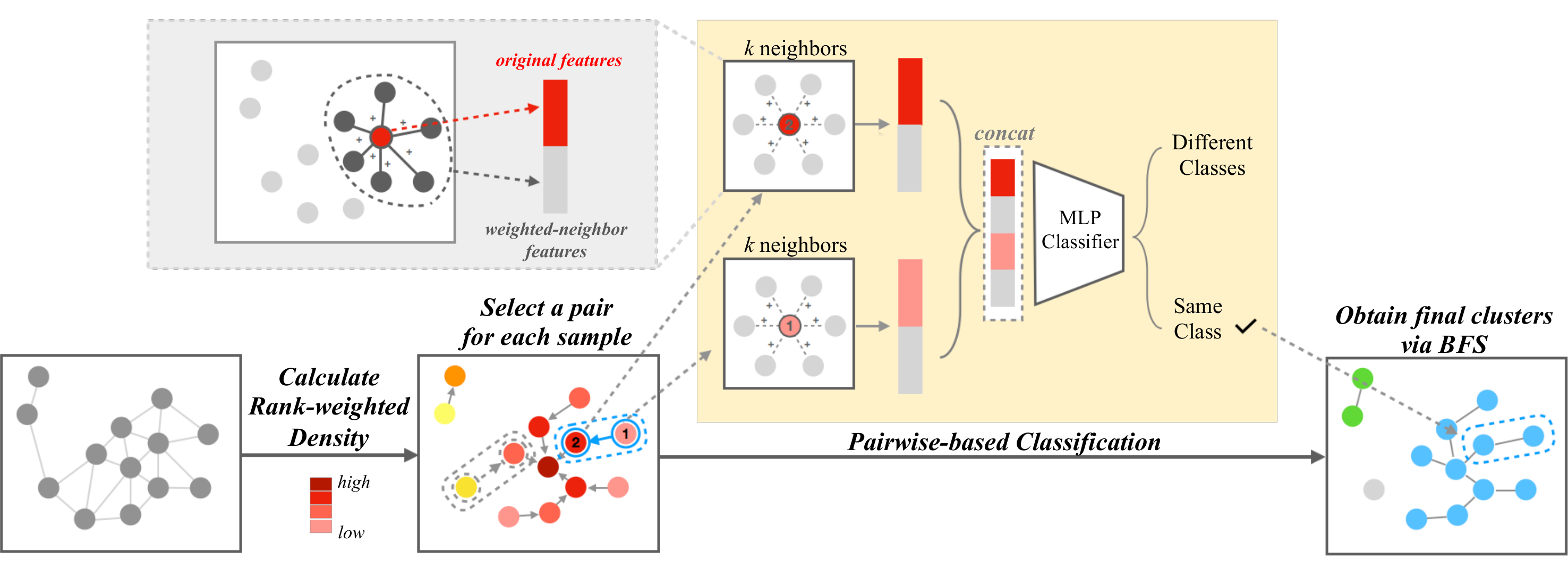}
\end{center}
   \caption{Overview of our proposed face clustering framework. We first use rank-weighted density to form pairs for each sample. The feature vector of each sample is updated by concatenating its original feature and weighted-neighbor feature, and we combine the updated features of a pair of samples to form the pair features, then each pair is sent to a trained classifier to determine the pairwise relationship. After the linkages of samples are determined, we apply BFS algorithm to obtain the final clusters.}
\label{fig1}
\end{figure*}
The challenges in face clustering mainly result from the large variations of face representations in the real world \cite{he2018merge}. Most existing graph-based clustering methods achieve satisfying performance by learning representative features from an affinity graph, yet relies on empirical thresholds to finally decide whether two samples should belong to the same class. Thus, performances of such methods are heavily influenced by manual settings and could drop severely in real-world face clustering scenes.

\noindent \textbf{Pairwise-relationship Prediction in Face Clustering.} Determining the pairwise relationship between samples is an essential part of the face clustering task in inference. Zhan {\em et al.} \cite{zhan2018consensus} aggregated multi-view information and discovered robust pairs by an MLP classifier, and finally set an empirical threshold to determine pairwise relationships between two samples. Wang {\em et al.} \cite{wang2019linkage} predicted linkage-probability between nodes and found connections with dynamic thresholds. Yang {\em et al.} \cite{yang2020learning} utilized a pre-defined threshold to cut off the edges with small similarities despite the introduced connectivity estimator. Guo {\em et al.} \cite{guo2020density} proposed the learnable local and long-range features to impair sensitivity to the clustering threshold, but a suitable threshold was still needed to decide the ultimate pairwise relationship. Different from previous approaches, our method focuses on exploiting a robust module to discover the pairwise relationship and aims at getting rid of the empirical threshold. This allows our method to be more generalizable in real-world face clustering scenes.

\noindent \textbf{GNN-based Face Clustering.} Recent works on face clustering task primarily adopt GNN-based models to learn representative feature embeddings, and tend to estimate relative order between samples by utilizing the ``density" \cite{{rodriguez2014clustering}, {ankerst1999optics}} or ``confidence" \cite{yang2020learning} of each sample, leading to state-of-the-art performances.  Wang {\em et al.} \cite{wang2019linkage} leveraged a graph convolutional network (GCN) model to find the connectivities between a pivot node and its neighbors.  Yang {\em et al.} \cite{yang2019learning} learned clusters in GCN detection and segmentation modules. In \cite{yang2020learning}, Yang {\em et al.} proposed an improved GCN-based approach to learn vertex confidence on the massive affinity graph and edge connectivity on the local subgraphs afterward. Guo {\em et al.} \cite{guo2020density} exploited cluster-level feature distribution by learning context-aware feature embeddings based on GCNs. The success of GNN-based methods mainly lies in the learning of context-aware features, where the neighborhood information is greatly emphasized. However, GNN-based methods can easily cause high computational cost, thus cannot work on large-scale clustering scenes on the condition of memory insufficiency. 

In this paper, we address the memory-consuming problem raised in popular state-of-the-art methods and free the face clustering methods from setting empirical thresholds by learning robust pairwise relationship, and further improves the accuracy and efficiency by introducing rank-weighted density.

\section{Methodology}
\label{sec3}
As has been discussed above, current GNN-based face clustering algorithms could be very memory-consuming, and rely on expert knowledge to set proper thresholds to determine the connectivities between faces. To address such challenges, we propose a face clustering method based on pairwise classification. We formulate the face clustering task as a classification task between faces and train a classifier to directly determine whether two faces should belong to the identical class. Furthermore, to generate face clusters more efficiently, we select face pairs sent to the classifier based on a novel rank-weighted density manner, which turns out to be more insensitive to outliers. Figure~\ref{fig1} illustrates an overview of our proposed method.

\subsection{Pairwise-based Classification}
Given a face dataset $F=[\bm{f_1},\bm{f_2},...,\bm{f_N}] \in \mathbb{R}^{N\times D}$, where $N$ is the number of face images and $D$ the dimension of features extracted from a trained CNN, face clustering is a task of dividing faces into different clusters where faces in the same cluster share one identity. Many popular face clustering methods focus on learning representative embeddings, then use empirical thresholds to determine the connectivities between samples, and such thresholds have great influences on the final performance of face clustering but rely heavily on experiences and vary greatly with different datasets. Meanwhile, the graph learning on large-scale graphs (\textit{e.g.}, with nodes more than 10 million) can be very memory-consuming. To resolve these problems, we change the learning from the {\em graph-level} to the {\em pair-level}. Given a pair of faces, what we look for is a binary classifier to directly give the relationship between these two faces, {\em i.e.}, whether they belong to the same identity or not.

\noindent \textbf{Motivation of classifier.} The huge memory consumption in graph learning-based methods results from the huge size $N$ of the dataset. But we do not necessarily need to learn knowledge on such a big scale and face clustering actually tends to assign samples with pseudo labels, so we solve the face clustering task on the minimum level, the pair-level, and train a classifier to predict whether two samples should belong to the same class. Take a pair of image features $\bm{f_1}, \bm{f_2}$ as input, naturally, we can concatenate the two feature vectors to form a new feature vector sent to the network, and we only need a network as simple as Multi-Layer Perceptron (MLP). One can also see that our method can be heuristically generalized to other kinds of classifiers.

\begin{algorithm}[t]
 \caption{Pairwise-based Clustering}
 \begin{algorithmic}[1]
 \renewcommand{\algorithmicrequire}{\textbf{Input:}}
 \renewcommand{\algorithmicensure}{\textbf{Output:}}
 \Require feature sets $\mathcal{F}$, the number of neighbors $k$, trained classifier $\mathcal{C}$.
 \Ensure clusters $\mathbb{C}$
    
\Procedure{Clustering}{}
\State $\mathbb{P}$ = \Call{Find Pairs via Density}{$\mathcal{F}$, $k$}
\State $\mathbb{R}$ = Use $\mathcal{C}$ to get relationships of pairs in $\mathbb{P}$.
\State $\mathbb{C}$ = Use BFS Algorithm to generate clusters.
\State \Return $\mathbb{C}$
\EndProcedure
\Statex
\Function{Find Pairs via Density}{$\mathcal{F}$, $k$}
\State $\mathbb{P}$ = $\emptyset$
\For {\textbf{all} sample $\textbf{\em x}$ in $\mathcal{F}$}
\State {Obtain similarities with its $k$ nearest neighbors.}
\State {Calculate its rank-weighted density.}
\EndFor
\For {\textbf{all} sample $\textbf{\em x}$ in $\mathcal{F}$}
\State {Find the first neighbor $\textbf{\em y}$ with higher density.}
\If {$\textbf{\em y}$ exists}
\State {$\mathbb{P}$ = $\mathbb{P}$ $\cup$ \{$(\textbf{\em x}, \textbf{\em y})$\}}
\EndIf
\EndFor
\State \Return $\mathbb{P}$
\EndFunction
 \end{algorithmic}
 \label{alg}
\end{algorithm}

Inspired from GNN-based face clustering methods which utilize contextual information to help cluster faces, and the intuition that a sample is very likely to share identity with its neighbors, we use neighbor features to refine its features. For sample ${x_{i}}$, we use \textit{k}-NN to select its top \textit{k} nearest neighbors ${x_{i1}}$, ${x_{i2}}$,..., ${x_{ik}}$, and similarities ${s_{i1}}$, $s_{i2}$,..., $s_{ik}$ between sample ${x_{i}}$ and its neighbors can be calculated as the inner product of two features. Instead of a simple summation, we apply the similarities as the weights to calculate the weighted-neighbor features of a sample:$$\bm{f_{i}^{'}}=\sum_{j=0}^{k}{s_{ij}\bm{f_{ij}}},\eqno(1)$$where $j=0$ indicates the feature of the sample itself and $s_{i0}$ is set to 1. This actually makes the intra-class features more compact and the inter-class samples more separable. However, considering samples spreading nearby the boundaries of different classes in feature space, the overladen attention to neighbors may bring errors to the classification, we thus concatenate the original features and the weighted-neighbor features to form the combined features, the pair features are finally determined as:$$\bm{F_{12}}=[\bm{f_{1}}, \bm{f_{1}^{'}}, \bm{f_{2}}, \bm{f_{2}^{'}}].\eqno(2)$$

Compared with graph-based methods, by learning on the pair-level, the memory usage is largely reduced in both the training and inference stages, and we can flexibly change the batch size to fit in the memory capacity available.

\noindent \textbf{Training set formulation.} 
For convenience, we define the pairs with the same class as positive pairs and pairs with different classes as negative ones. To formulate the training set, for each face, we collect the set of faces of the same class and combine all these sets as the positive samples. For negative samples, we use \textit{k}-NN to find the top \textit{k} nearest neighbors of a face. Then we choose the neighbors with different classes as negative samples, which are naturally hard negative samples, for they are close enough in the feature space but belong to different classes. We vary the $k$ value until the number of negative samples is almost equal to positive ones in the training set.
\subsection{Rank-weighted Density}
\begin{figure}[t]
\begin{center}
   \includegraphics[width=1.0\linewidth]{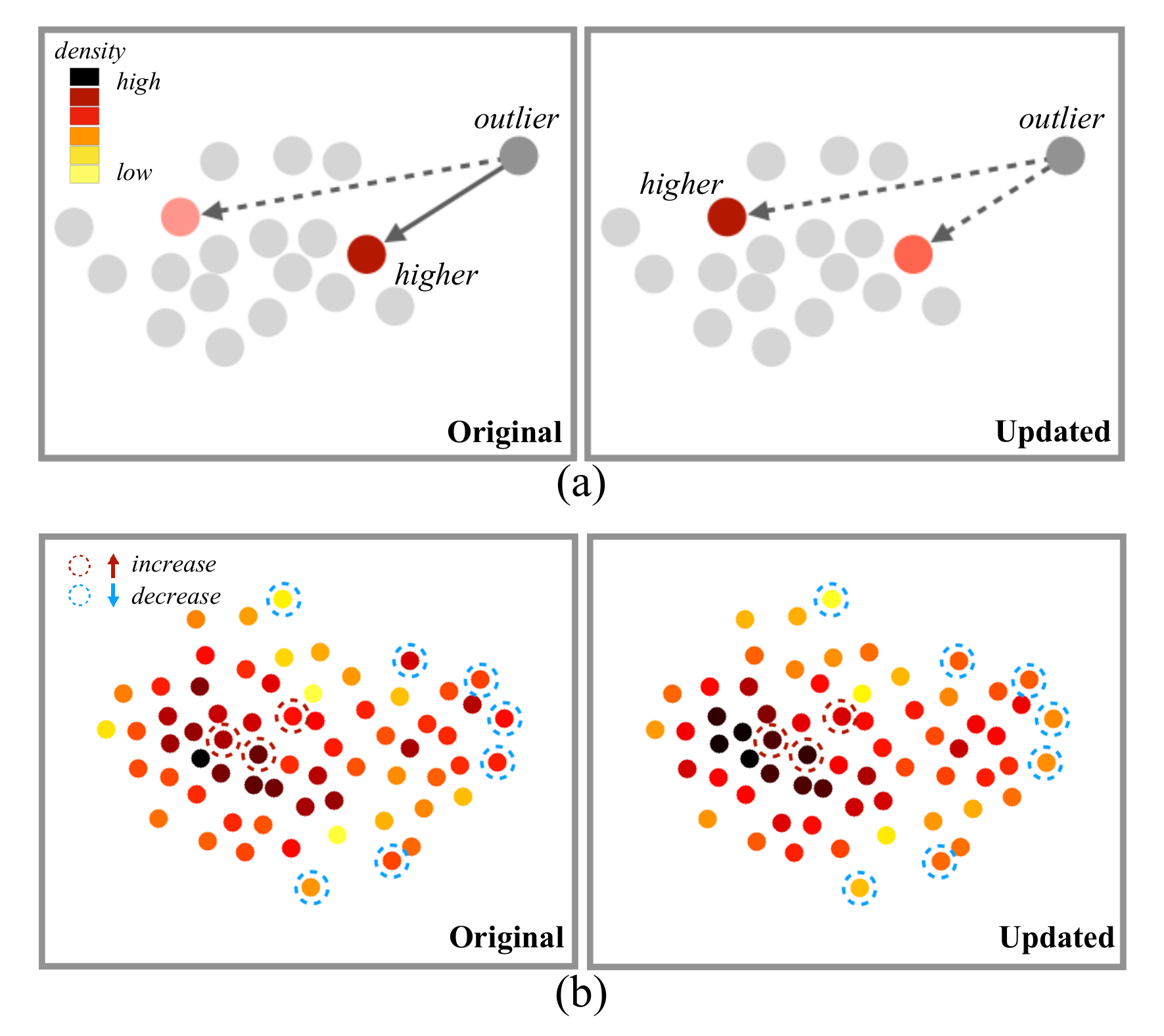}
\end{center}
   \caption{The visualization of how our proposed rank-weighted density outperforms the original density. (a) The updated rank-weighted density can ease the cluster-center-shift caused by outliers. (b) Feature distribution of a cluster from the MS-Celeb-1M dataset \cite{guo2016ms} on t-SNE. The relative orders of densities of border samples decrease and those of centroid samples increase.}
\label{fig2}
\end{figure}
It is time-consuming to predict the relationships of all pairs in the inference stage, for a test set of size $N$ faces will result in a test pair set of size $N^{2}$, and inference results of such big test pairs set will cause insufficiency in the breadth-first search (BFS) process to obtain the final clusters. To enhance the efficiency of the clustering pipeline, we use density to select appropriate pairs, thus reducing the number of pairs sent to the classifier.

Intuitively, in a cluster, the {\em border} samples can be not quite similar to each other, but the {\em centroid} samples are very likely to be similar to almost all samples. So we try to pair a sample with the {\em centroid} sample. The original ``density''\cite{{ester1996density},{ankerst1999optics},{guo2020density}} of a sample can measure how likely a sample to be cluster center or how close a sample is from the cluster center, and is defined as the sum of the similarities of the top {\em k} nearest samples with the target sample. The calculation of similarity has been introduced previously. For instance, given a sample, the similarities between its neighbors and itself can be calculated, and the density $d_{i}$ of sample $x_{i}$ is defined as:$$d_{i}=\sum_{j=1}^{k}s_{ij},\eqno(3)$$where \textit{k} is the number of neighbors.

However, due to the existence of outliers which begets that samples may get a higher density but further to the cluster center compared to another sample, the original density can cause inferior density orders and generate inappropriate pairs. And the outliers can influence the density of samples even more greatly in dense cluster distribution. To depress the adverse effects of outliers, considering that they are often relatively distant neighbors of most samples, we update density in such manner:$$d_{i}^{'}=\sum_{j=1}^{k}f(j)s_{ij},\eqno(4)$$where $f(j)$ is a monotonically decreasing function, which indicates that the closer neighbors should weight more in the calculation of density. When our updated rank-weighted density disagrees with the original density, for instance, sample ${x_{i}}$ has a higher updated density compared to sample ${x_{j}}$ while lower in original density, that is because the samples not close with the target sample ${x_{j}}$ are given less attention despite that they have high similarities with ${x_{j}}$. Figure~\ref{fig2} (a) illustrates how our updated density outperforms the original density. In original density, although ${x_{i}}$ tends to be in a more dense situation and more likely to be close to the cluster center, the outlier helps ${x_{j}}$ gain a higher density, thus cause the \textbf{\em cluster-center-shift} among samples. We present the feature distribution of a cluster sampled from the MS-Celeb-1M dataset \cite{guo2016ms} in Figure~\ref{fig2} (b). It can be clearly seen that the densities of both {\em border} samples and {\em centroid} samples are optimized. With our novel rank-weighted density, the likelihood of a sample being cluster center is more precise and less insensitive to outliers.

We now use our updated rank-weighted density to instruct the selection of pairs sent to the classifier in inference. For each sample, we select the first sample with a higher density among its \textit{k} nearest neighbors and form a pair. The selected sample is closer to the cluster center yet still similar enough to share the same class with the target sample. If a sample has no neighbors with higher density, no neighbors would be selected. But it can be seen that such samples are very likely to be picked by other samples and form pairs. And we only need to test size of no more than $N$ pairs sent to the classifier, which largely increases the efficiency in inference.

\begin{table} []
\centering
\setlength{\tabcolsep}{4mm}
\begin{tabular}{c|cc|c}
\hline
Methods  & $F_P$ & $F_B$ & Time \\\hline\hline
K-Means \cite{{lloyd1982least},{sculley2010web}} &79.21 & 81.23 & 11.5h \\
HAC \cite{sibson1973slink} & 70.63 & 70.46 & 12.7h \\
DBSCAN \cite{ester1996density} & 67.93 & 67.17 & 1.9m \\
ARO \cite{otto2018clustering} & 13.6 & 17  & 27.5m \\
CDP \cite{zhan2018consensus} & 75.02& 78.7 & 2.3m \\
L-GCN \cite{wang2019linkage} &78.68 & 84.37 & 86.8m \\
LTC \cite{yang2019learning} & 85.66 & 85.52 & 62.2m \\
GCN(V+E) \cite{yang2020learning} & 87.93 & 86.09 & 11.5m \\
DANet \cite{guo2020density} & 90.6  & $-$ & 5.5m \\\hline\hline
\textbf{Ours}    &  \textbf{90.67} & \textbf{89.54} & \textbf{1.7m} \\ \hline
\end{tabular}
\caption{Comparison on MS-Celeb-1M.}
\label{tab0}
\end{table}

\subsection{Complexity Analysis}
The time complexity of our method mainly arises from the $k$-NN search. Thanks to the approximate nearest search proposed in \cite{wieschollek2016efficient}, the time complexity is reduced to $\mathcal{O}(nlogn)$ and can be further accelerated by GPU \cite{johnson2019billion}. Once the $k$-NN search is finished, as shown in Algorithm~\ref{alg} the density calculation of each sample, the pair selection and the feature construction of all test samples are all $\mathcal{O}(n)$. The number of pairs selected sent to the classifier is no more than the number of the samples, so the time cost is also $\mathcal{O}(n)$. Consequently, the overall time complexity of our method is $\mathcal{O}(nlogn)$, and the time consumption mainly lies in the $k$-NN search.

Our method is also memory-friendly. The popular graph learning-based method takes $\mathcal{O}(ND)$ space complexity to learn on the whole graph, where $N$ is the number of samples which can be up to millions and $D$ the length of features. Commonly, $N$$\gg$$D$. In our proposed method, the inputs of the network are only concatenated feature vectors with lengths at $D$, and the large number $N$ is divided into batches in training or inference, so our method reduces memory consumption by a very considerable margin. The details of our time and memory consumption will be revealed in Sec.~\ref{sec4}.
\section{Experiments}
\label{sec4}
\subsection{Experimental Settings}
\noindent \textbf{Datasets.} We first evaluate the proposed method on two large public face clustering benchmarks, MS-Celb-1M\cite{guo2016ms} and IJB-B\cite{whitelam2017iarpa}. Following the widely used annotations from ArcFace \cite{deng2019arcface}, a reliable subset containing 5.8{\em M} images from 86{\em K} ids of MS-Cleb-1M is collected. The training and testing sets are divided following the settings as \cite{yang2020learning}, where the subset was split into 10 parts almost equally. We train on one labeled part and select 1, 3, 5, 7 and all parts from the other 9 unlabeled parts, resulting in test sets with sizes of 584{\em K}, 1.74{\em M}, 2.89{\em M}, 4.05{\em M} and 5.21{\em M} images respectively. For IJB-B, we use the same settings as in \cite{wang2019linkage}, adopting a random subset of the CASIA dataset \cite{yi2014scratch} which contains 5{\em K} ids and 200{\em K} samples as training set, and test on the three largest subtasks of IJB-B. The number of ids in the three subtasks are 512, 1,024 and 1,845, containing 18,171, 36,575 and 68,195 images respectively. We also evaluate our clustering approach on datasets other than face images to prove the generalization of our proposed method on clustering tasks. For a fair comparison, we use the same settings as in \cite{yang2020learning}, where the training set with around 26{\em K} images from 4{\em K} categories and test set with around 27{\em K} images from 4{\em K} categories are sampled from the DeepFashion \cite{liu16deepfashion} benchmark.

\noindent \textbf{Metrics.} The performances of our proposed method and comparison methods are evaluated on the two popular clustering metrics \cite{shi2018face}, namely Pairwise F-score and BCubed F-score \cite{amigo2009comparison} respectively. Both metrics are calculated as the harmonic mean of precision and recall, and will be referred to as $F_P$ and $F_B$ in the following sections. 
\begin{figure}[t]
\begin{center}
   \includegraphics[width=1\linewidth]{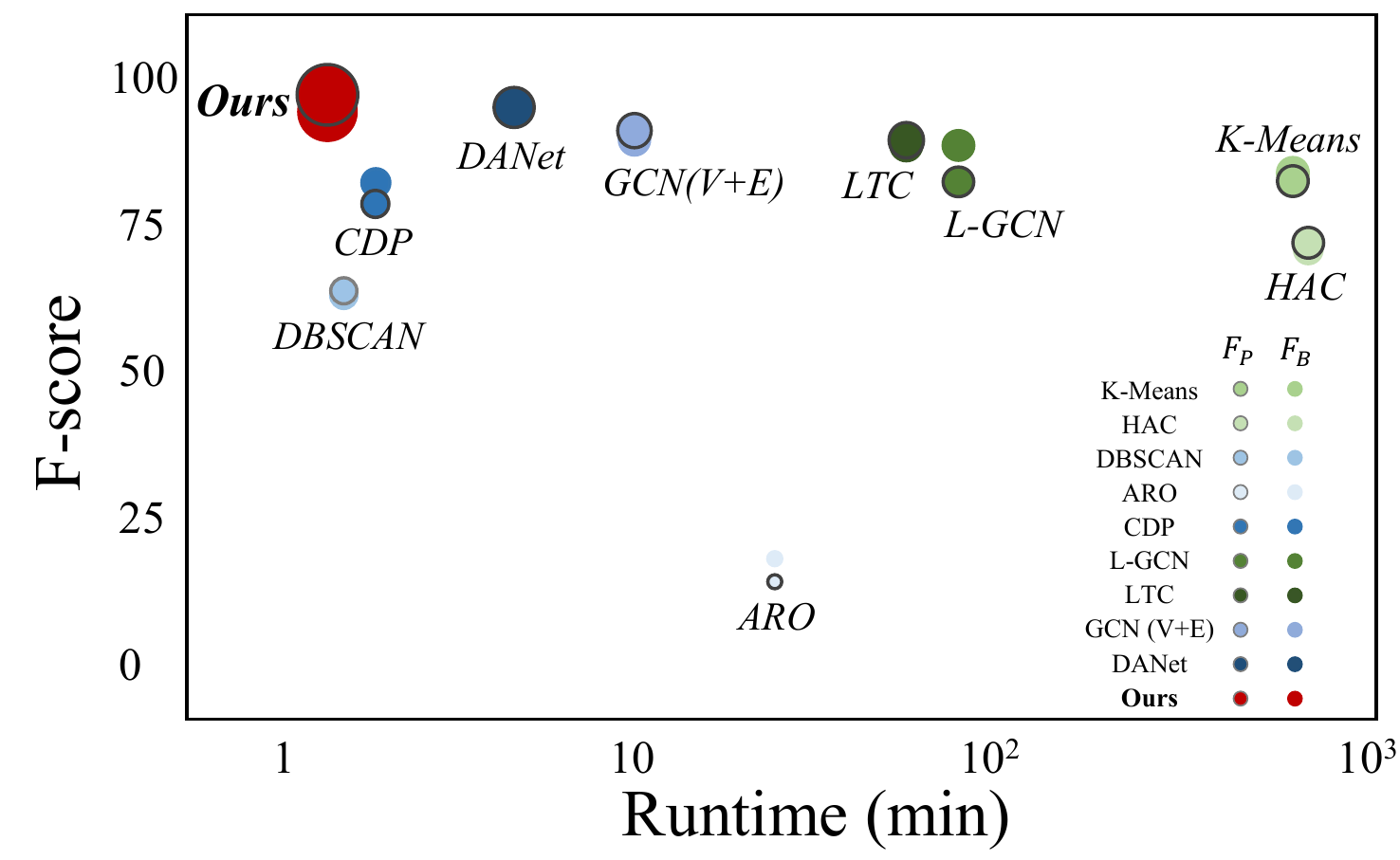}
\end{center}
   \caption{The trade-off between accuracy and efficiency of our method and comparison methods. Note that the runtime axis is in log-scale.}
\label{fig3}
\end{figure}

\begin{table*} []
	\centering
	
	\begin{tabular}{c|cc|cc|cc|cc|cc}
		\hline
		Number of images& \multicolumn{2}{c|}{584K} & \multicolumn{2}{c|}{1.74M} &\multicolumn{2}{c|}{2.89M} & \multicolumn{2}{c|}{4.05M} &\multicolumn{2}{c}{5.21M} \\\cline{1-11}
		Methods$/$ Metrics  & $F_P$ & $F_B$ & $F_P$ & $F_B$ & $F_P$ & $F_B$ 
		& $F_P$ & $F_B$ & $F_P$ & $F_B$ \\\hline\hline
		K-Means \cite{{lloyd1982least},{sculley2010web}} & 79.21 & 81.23 & 73.04 &
		75.2 & 69.83 & 72.34 & 67.9 & 70.57 & 66.47 & 69.42 \\
		HAC \cite{sibson1973slink} & 70.63 & 70.46 & 54.4 & 69.53 & 11.08 & 68.62 &
		1.4 & 67.69 & 0.37 & 66.96 \\ 
		DBSCAN \cite{ester1996density} & 67.93 & 67.17 & 63.41 & 66.53
		& 52.5 & 66.26 & 45.24 & 44.87 & 44.94 & 44.74 \\ 
		ARO \cite{otto2018clustering} & 13.6 & 17 & 8.78 & 12.42 & 7.3 & 10.96
		& 6.86 & 10.5 & 6.35 & 10.01 \\ 
		CDP \cite{zhan2018consensus} & 75.02 & 78.7 & 70.75 & 75.82 & 69.51
		& 74.58 & 68.62 & 73.62 & 68.06 & 72.92 \\ 
		L-GCN \cite{wang2019linkage} & 78.68 & 84.37 & 75.83 & 81.61 & 74.29 &
		80.11 & 73.7 & 79.33 & 72.99 & 78.6 \\ 
		LTC \cite{yang2019learning} & 85.66 & 85.52 & 82.41 & 83.01 & 80.32 &
		81.1 & 78.98 & 79.84 & 77.87 & 78.86 \\ 
		GCN(V+E) \cite{yang2020learning} & 87.93 & 86.09 & 84.04 & 82.84 & 82.1 & 81.24 &
		80.45 & 80.09 & 79.3 & 79.25 \\\hline\hline
		\textbf{Ours}    & \textbf{90.67} & \textbf{89.54} & \textbf{86.91} & \textbf{86.25} & \textbf{85.06} & \textbf{84.55} & \textbf{83.51} & \textbf{83.49} & \textbf{82.41} & \textbf{82.4} \\ \hline
	\end{tabular}
	\caption{Comparison on face clustering with different numbers of unlabeled images from the MS-Celeb-1M dataset.}
	\label{tab1}
\end{table*}

\noindent \textbf{Implementation Details.} To construct the neighbor features in pairwise relationship classifier and calculate the rank-weighted density, for each dataset, we use the same $k$ in all $k$-NN scenes. The choice of $k$ mainly depends on the sizes of clusters in the training set. Specifically, the $k$ is set to 80, 40 and 5 for MS-Celeb-1M, IJB-B and DeepFashion respectively. The classifier is stacked with 3 fully connected layers. In training, we use SGD optimizer with start learning rate 0.01, momentum 0.9 and weight decay 5e-4. The batch size is set to 2048 and the training stops after 60 epochs. In the calculation of rank-weighted density, we use the power function as the weighting function:$$f_{i}=(k-i)^p.\eqno(5)$$The setting of $p$ will be further discussed in ablation studies.

\subsection{Method Comparison}
We compare our method with a series of face clustering baselines. These approaches can be generally categorized to conventional methods and learning based methods. The representative approaches in the former category include K-means \cite{{lloyd1982least},{sculley2010web}}, Hierarchical Agglomerative Clustering (HAC) \cite{{sibson1973slink}, {mullner2013fastcluster}}, Density-Based Spatial Clustering of Applications with Noise (DBSCAN) \cite{ester1996density}, and Approximate Rank Order (ARO) \cite{otto2018clustering}. The latter shows more promising results, including Consensus-Driven Propagation (CDP) \cite{zhan2018consensus}, L-GCN \cite{wang2019linkage}, Learning to Cluster (LTC) \cite{yang2019learning}, GCN (V+E) \cite{yang2020learning} and DANet \cite{guo2020density}, where the last two achieve the previously best performances on $F_B$ and $F_P$ respectively.

\begin{table} []
\setlength{\tabcolsep}{3.6mm}
	\centering
	\begin{tabular}{cccc}
		\hline
		Methods  & $F_{512}$ &  $F_{1024}$ & $F_{1845}$  \\\hline\hline
		K-Means \cite{lloyd1982least} & 61.2 & 60.3 & 60.0 \\
		DBSCAN \cite{ester1996density} & 75.3 & 72.5 & 69.5 \\  
		ARO \cite{otto2018clustering} & 76.3 & 75.8 & 75.5 \\ 
		L-GCN \cite{wang2019linkage} & 83.3 & 83.3 & 81.4 \\
		DANet \cite{guo2020density} & 83.4  & 83.3 & \textbf{82.8} \\ \hline\hline
		\textbf{Ours}  & \textbf{84.4} & \textbf{83.3} & 82.7 \\ \hline
	\end{tabular}
    \caption{Comparison on IJB-B. $F_{512}$, $F_{1024}$ and $F_{1845}$ are Pairwise F-scores of different test sets.} 
    \label{tab2}
\end{table}

\begin{table} []
	\centering
	\begin{tabular}{ccccc}
		\hline
		Methods  & Clusters &  $F_P$ & $F_B$  & Time \\\hline\hline
		K-Means \cite{lloyd1982least} & 3991 & 32.86 & 53.77 & 573s \\
		HAC \cite{sibson1973slink} & 17410 & 22.54 & 48.77 & 112s \\ 
		DBSCAN \cite{ester1996density} & 14350 & 25.07 & 53.23 & 2.2s \\ 
		MeanShift \cite{cheng1995mean} & 8435 & 31.61 & 56.73 & 2.2h \\ 
		Spectral \cite{ho2003clustering} & 2504 & 29.02 & 46.4 & 2.1h \\ 
		ARO \cite{otto2018clustering} & 10504 & 26.03 & 53.01 & 6.7s \\ 
		CDP \cite{zhan2018consensus} & 6622 & 28.28 &57.83 & 1.3s \\ 
		L-GCN \cite{wang2019linkage} & 10137 & 28.85 & 58.91 & 23.3s \\ 
		LTC \cite{yang2019learning} & 9246 & 29.14 & 59.11 & 13.1s \\ 
		GCN(V+E) \cite{yang2020learning} & 6079 & \textbf{38.47} & 60.06 & 18.5s \\ \hline\hline
		\textbf{Ours}    & \textbf{6018} & 37.67 & \textbf{62.17} & \textbf{0.6s} \\ \hline
	\end{tabular}
    \caption{Comparison on DeepFashion.}
    \label{tab3}
\end{table}

\begin{figure}[t]
\begin{center}
   \includegraphics[width=0.8\linewidth]{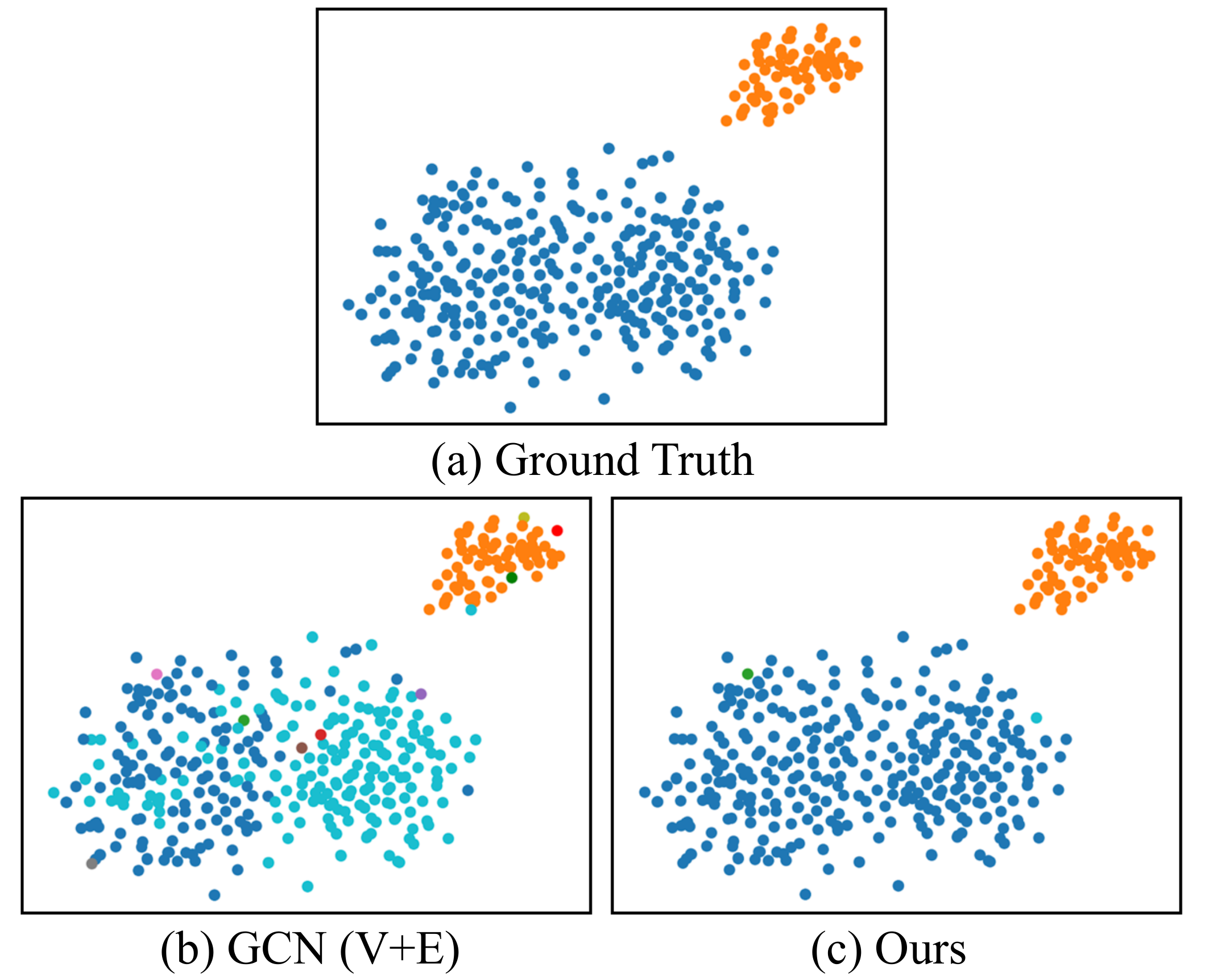}
\end{center}
   \caption{The feature distribution of two clusters of samples from the MS-Celeb-1M dataset on t-SNE. (a) The ground truth. Nodes with the same colors belong to the same class. (b) Result of GCN (V+E) \cite{yang2020learning}. (c) Result of our method. Our method predicts more accurately and generates fewer singleton clusters.}
\label{fig6}
\end{figure}

\noindent \textbf{Results.} We firstly test our method and other comparison methods on the first test part of the MS-Celeb-1M data, containing 580{\em K} images of 8,573 identities as in \cite{{zhan2018consensus}, {guo2020density}}. The results presented in Table~\ref{tab0} compares the Pairwise F-score and BCubed F-score performances of different approaches on this set. Our method achieves state-of-the-art performance in terms of both metrics, surpassing all previous methods. Figure~\ref{fig6} visualizes two clusters of the test set on t-SNE, where we can see that our method performs better with generating fewer singleton clusters and clustering samples from a large cluster more accurately. By formulating face clustering as pairwise classification and learning on pair-level instead of graph-level, our method actually handles the face clustering task from a fundamental perspective and faces the ultimate question of clustering, {\em i.e.}, the homogeneity of several samples.

\begin{figure}[t]
\begin{center}
   \includegraphics[width=0.9\linewidth]{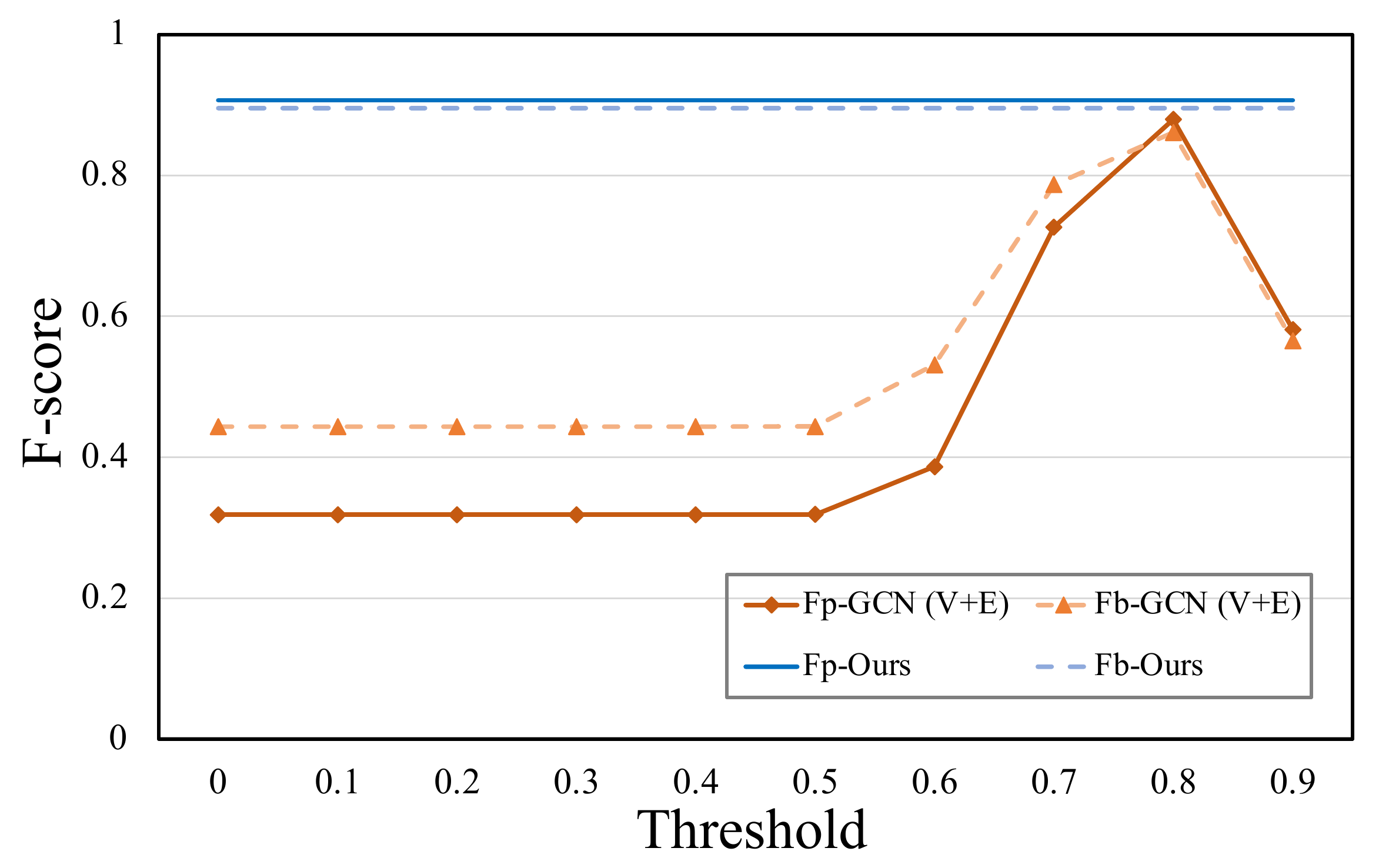}
\end{center}
   \caption{The performances of GCN (V+E) \cite{yang2020learning} under different thresholds in inference on MS-Celeb-1M dataset. The two lines on top are the $F_P$ and $F_B$ performances of our method.}
\label{fig5}
\end{figure}

To demonstrate the robustness of our threshold-free method, the $F_P$ and $F_B$ performances of the state-of-the-art method GCN (V+E) \cite{yang2020learning} under different thresholds in inference are assessed. Figure~\ref{fig5} shows that performances of this method are quite sensitive to the threshold, which introduces inconsistency when applied to various real-world scenes. But our method demands no threshold and thus is a more robust method.

In addition, we perform an experiment that compares the generalization of different methods on incremental numbers of unlabeled images. The results in Table~\ref{tab1} shows that our method can generalize better and achieve consistent improvements on larger test sets compared with other methods. It demonstrates that our proposed method can obtain more robust and excellent performances in large-scale scenes. Table~\ref{tab2} gives the results of our method and comparison methods on another large face clustering benchmark IJB-B. It can be seen that our method gains comparative or even better performances on the three largest subtasks of this dataset.

Table~\ref{tab3} shows that in other clustering benchmarks like DeepFashion, our method also gains satisfying performances by outperforming state-of-the-art methods in BCubed F-score with a 2.11\% lead, which further reveals the generalization ability of our proposed method on clustering tasks.

\noindent \textbf{Memory and Time Consumption.} We run our experiments on a single Telsa P40 with 24G memory capacity. As has been discussed above, the graph learning on large-scale graphs could be very memory-consuming, thus become impractical in certain real-world scenes. Take GCN (V+E) \cite{yang2020learning} for example, to reduce memory consumption, GCN-V only uses one layer but still raises an out-of-memory error when the size of test set comes to 2.89{\em M} on MS-Celeb-1M dataset in inference. Yet in our approach, since the inputs of the classifier are only 1-D features with batch input, we can apply our method in any size of datasets and flexibly set the batch size to fit our accessible calculation resources, even only on CPU. And the batch size of 2048 only needs 0.7G memory. 

From Table~\ref{tab0}, our method is also time-efficient, and actually the fastest one, even surpassing the conventional methods. For a fair comparison, we analyze the inference time of all supervised methods on MS-Celeb-1M with $N=584K$ as in \cite{yang2020learning}, and our method takes about 1.7{\em m} on a single GPU with batch size 2048. Figure~\ref{fig3} illustrates the trade-off between accuracy and efficiency. It shows that our method not only achieves state-of-the-art performance, but also enjoys a large advantage on time consumption. To conclude, our method achieves state-of-the-art performance at the fastest speed with very limited memory usage.

\subsection{Ablation Study}
We conduct the ablation study mainly on the MS-Celeb-1M dataset.

\noindent \textbf{Design of input features.} We explore different designs of input features sent to the classifier. As has been discussed in Sec.~\ref{sec3}, given two features, the first thought would be a simple concatenation of the two original feature vectors, which we note as {\em original features}. To emphasize the importance of neighbor features, we design the similarity-weighted summation of neighbor features and note them as {\em weighted-neighbor features}. The concatenation of the original features and the weighted neighbor features are finally noted as {\em combined features}. We use exactly the same hyper-parameters to train the classifier based on different kinds of features for a clearer comparison.
\begin{table}
\centering
\begin{tabular}{cccc}
\hline
Feature Setting  & Precision & Recall & Accuracy \\\hline\hline
original  & 92.2 & 83.2 & 84.7  \\
weighted-neighbor  & 94.4 & \textbf{97.9} & 95.2 \\
\textbf{combined}    & \textbf{95.4} & 97.5 & \textbf{95.6} \\ \hline
\end{tabular}
\caption{Performances of our classifier based on different features. The positives are sample pairs with the same class.}
\label{tab4}
\end{table}

\begin{table}
\centering
\setlength{\tabcolsep}{1.8mm}
\begin{tabular}{ccccccc}
\hline
Power  &  $Pre_P$ & $Rec_P$ & $F_P$ & $Pre_B$ & $Rec_B$ & $F_B$\\\hline\hline
0  & 88.04 & 89.1 & 88.57 & 91.63 & 85.19 & 88.3\\
0.5  & 89.16 & 89.12 & 89.14 & 92.33 & 85.37 & 88.71\\
1  & 90.26 & \textbf{89.13} & 89.69 & 92.91 & \textbf{85.5} & 89.05\\  
3  & 92.11 & 88.88 & 90.47 & 93.84 & 85.44 & 89.45\\ \hline
\textbf{5}  & \textbf{92.94} & 88.5 & \textbf{90.67} & \textbf{94.51} & 85.06 & \textbf{89.54}\\ \hline
7  & 92.84 & 87.93 & 90.32& 94.33 & 84.72 & 89.27\\ \hline
\end{tabular}
\caption{The influence of power on the MS-Celeb-1M dataset.}
\label{tab5}
\end{table}

As shown in Table~\ref{tab4}, the performance of the classifier based on weighted-neighbor features largely outperforms that on original features, for the nearby neighbors are more likely to share the same identity with the sample, thus providing contextual information on categories, which finally boosts the performance of the classifier. The combined features show a further improvement than weighted-neighbor features, for it gives more attention to the sample itself, which may reduce the possible influence of neighbors with different classes.

\begin{figure}[t]
\begin{center}
   \includegraphics[width=0.9\linewidth]{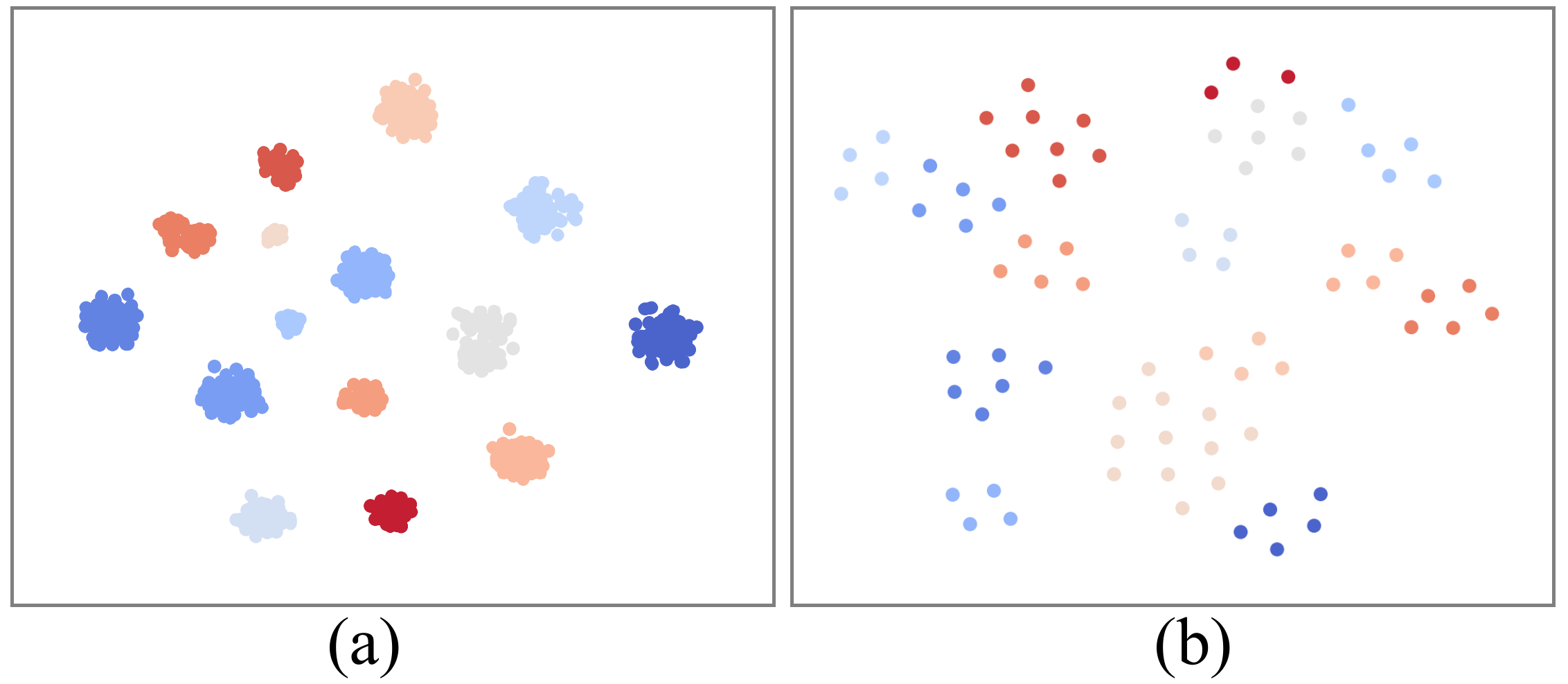}
\end{center}
   \caption{The feature distribution from different datasets on t-SNE. (a) The MS-Celeb-1M dataset. (b) The DeepFashion dataset.}
\label{fig4}
\end{figure}

\noindent \textbf{Design of rank-weighted density.} To give more attention to nearer neighbors in the calculation of density and reduce the influence of outliers, a monotonically decreasing function is applied to add weights on the summation of similarities. We choose the simple power function $f(i)=(k-i)^p$ and vary the power to obtain the best results.
Table~\ref{tab5} shows the influence of power on the MS-Celeb-1M dataset. The rank-weighted density turns into original density when power is set to 0. We can see that the designed rank-weighted density brings consistent performance gains and power 5 achieves the best balance between precision and recall. We believe the best choice of power on each dataset depends on the sparsity of the dataset itself. Figure~\ref{fig4} makes visualization of feature distribution from different datasets on t-SNE. On datasets like DeepFashion, the samples are distributed more evenly, and the influence of different choices of power is very limited. However, on large-scale datasets with more than 1{\em M} images like MS-Celeb-1M, the samples distribute very densely in feature space, the power should be higher to make the influences of neighbors with different distance ranks more distinguishable.

\section{Conclusion}
\label{sec5}
This paper has  proposed a simple yet elegant face clustering framework based on pairwise classification. We adopt a classifier to determine the relationships between samples, which largely reduce the memory consumption by learning on the pair-level instead of graph-level, and also frees face clustering task from the manual setting of thresholds in inference. Besides, to further enhance the efficiency, we design a novel rank-weighted density to guide the selection of pairs sent to the classifier. Extensive experimental results on public benchmarks demonstrate that our method achieves excellent performances in terms of both accuracy and efficiency, and also generalizes well in larger test sets and other clustering tasks.


\end{document}